\pdfoutput=1

\documentclass[11pt]{article}

\usepackage[preprint]{acl}

\usepackage{times}
\usepackage{latexsym}
\usepackage{multicol}
\usepackage{multirow, makecell, caption}
\usepackage{colortbl}
\usepackage{tikz}
\usepackage{arydshln}
\usepackage{pgfplots}
\usepackage{amsmath}
\usepackage{amssymb}
\usepackage{graphicx}
\usepackage{tabularx}
\usepackage{enumerate}
\usepackage{tcolorbox}
\usepackage{arydshln}
\usepackage{booktabs}
\usepackage{inconsolata}
\usepackage{microtype}
\usepackage{xcolor}
\usepackage{algorithm}
\usepackage{algorithmic}
\usepackage[utf8]{inputenc}
\usepackage{microtype}
\usepackage{inconsolata}
\usepackage{graphicx}
\usepackage{pifont}
\usepackage{xspace}
\usepackage{tcolorbox} 
\tcbuselibrary{listingsutf8}
\usepackage{listings}
\usepackage{hyperref}

\newcommand{\our}{UltraGen\xspace}

\title{\our: Extremely Fine-grained Controllable Generation via\\Attribute Reconstruction and Global Preference Optimization}

\author{Longfei Yun \and Letian Peng \and Jingbo Shang\thanks{$\ $  Corresponding author. } \\
University of California, San Diego \\
  \texttt{\{loyun, lepeng, jshang\}@ucsd.edu}
  }

\begin{document}
\maketitle
\begin{abstract}
Fine granularity is an essential requirement for controllable text generation, which has seen rapid growth with the ability of LLMs.
However, existing methods focus mainly on a small set of attributes like 3 to 5, and their performance degrades significantly when the number of attributes increases to the next order of magnitude.
To address this challenge, we propose a novel zero-shot approach for extremely fine-grained controllable generation (EFCG), proposing \emph{auto-reconstruction (AR)} and \emph{global preference optimization (GPO)}.
In the AR phase, we leverage LLMs to extract soft attributes (e.g., \textit{Emphasis on simplicity and minimalism in design}) from raw texts, and combine them with programmatically derived hard attributes (e.g., \textit{The text should be between 300 and 400 words}) to construct massive (around $45$) multi-attribute requirements, which guide the fine-grained text reconstruction process under weak supervision.
In the GPO phase, we apply direct preference optimization (DPO) to refine text generation under diverse attribute combinations, enabling efficient exploration of the global combination space. 
Additionally, we introduce an efficient attribute sampling strategy to identify and correct potentially erroneous attributes, further improving global 
optimization.
Our framework significantly improves the constraint satisfaction rate (CSR) and text quality for EFCG by mitigating position bias and alleviating attention dilution.\footnote{Code: \href{https://github.com/LongfeiYun17/UltraGen}{https://github.com/LongfeiYun17/UltraGen}}
\end{abstract}

\section{Introduction}
\begin{figure}[t] 
    \centering
        \includegraphics[width=0.48\textwidth]{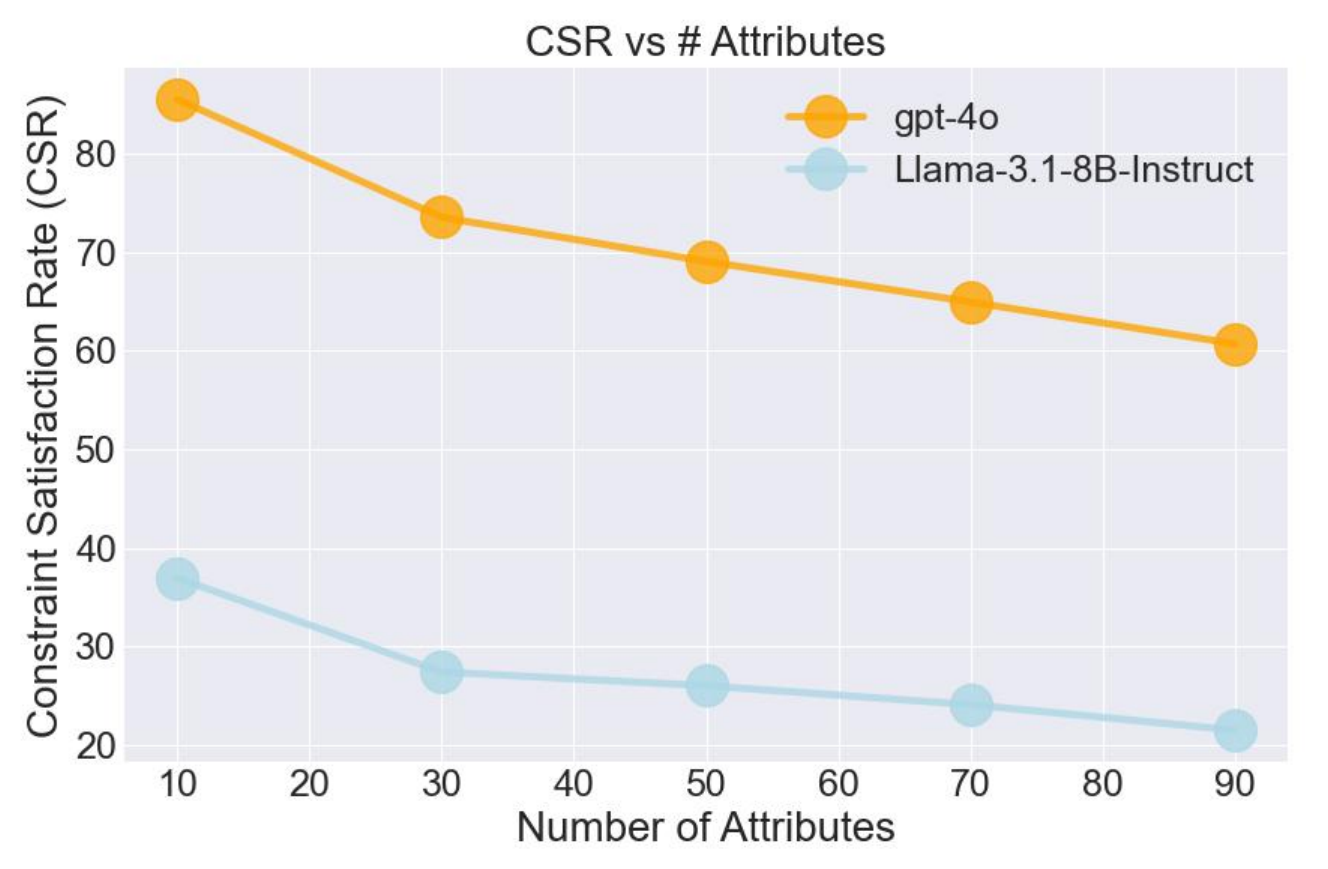}
    \caption{Constraint Satisfaction Rate (CSR) across different numbers of attributes for GPT-4o and LLaMA-3.1-8B-Instruct.}
    \vspace{-1.5em}
    \label{fig:teaser}
\end{figure}

While large language models (LLMs) have shown promising performance in various tasks, processing massive input information remains a challenging setup~\citep{lost-in-the-middle}.
For controlled text generation (CTG) ~\cite{wang2023aligning,song2024preference}, models are often required to satisfy a certain number of constraints simultaneously. In previous works, the typical number of constraints ranges from 3 to 5.
However, when the number of constraints scales to the extreme (e.g. 30 or more), performance degrades significantly (Figure ~\ref{fig:teaser}).

One representative challenge arises in travel itinerary planning (Appendix \ref{appendix:case_study}). Consider a prompt requiring a detailed 5-day travel plan that satisfies over 30 constraints, covering timing, budget, transportation, meal preferences, and specific landmark visits. Despite LLMs’ impressive fluency, they frequently violate crucial attributes like \textit{each activity must be under 2 hours} or \textit{avoid scheduling during 1 PM to 2 PM due to lunch break}. Why do LLMs perform well in general text generation but struggle under EFCG? We hypothesize that this limitation stems from two fundamental issues.

\begin{figure*}[t]
    \centering
        \includegraphics[width=1.0\textwidth]{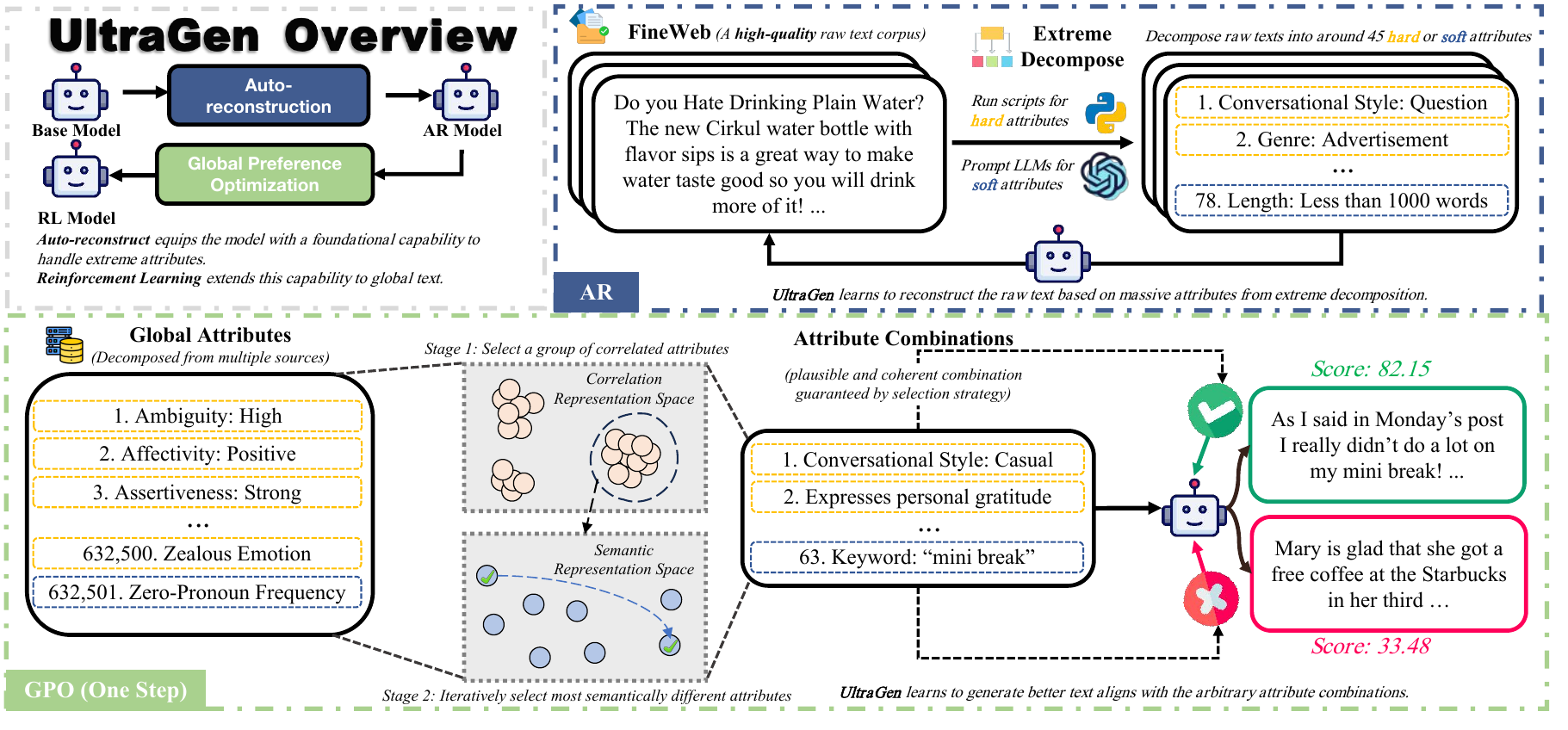}
    \caption{The whole pipeline of our two-stage \textbf{UltraGen} framework. The auto-reconstruction stage constructs a large-scale dataset by extracting soft and hard attributes from web corpora and then reconstructing the raw text. The global preference optimization stage applies DPO with attribute correlation modeling and diversity selection to enhance multi-attribute generalization over a global corpus.
    }
    \label{fig:pipeline}
    \vspace{-0.5em}
\end{figure*}

First, When the number of constraints increases, later-specified conditions are more likely to be neglected, as the model’s performance exhibits position-dependent degradation~\citep{lost-in-the-middle}, meaning that attributes appearing later in the prompt are less likely to be satisfied (Figure~\ref{fig:position_bias}). Second, the pretraining and instruction-tuning corpus of LLMs involve simple prompts with a few loosely defined requirements. Models rarely encounter instances with 30+ precise constraints in a single prompt. As a result, when faced with extreme constraint setups, models fail to maintain attention across all conditions, leading to attention dilution (Figure~\ref{fig:case_study}) during generation.

To tackle the challenges, we propose a framework designed to enhance LLMs’ ability to handle a massive number of constraints effectively. 
We hypothesize that position bias arises partly due to the lack of exposure to diverse attribute positions during training. To mitigate this, we construct a large-scale automated dataset pipeline that extracts soft attributes (e.g., style, content) and hard attributes (e.g., keywords, structure) from natural texts without human annotation. A quality check is conducted to ensure that the attributes align with the raw texts (Section ~\ref{sec:dataset_construction}). By training LLMs on these realistic multi-attribute inputs, we expose the model to variable attributes across different positions, enabling it to better internalize the relationship between attributes and the text regardless of position.

Besides, as the number of constraints increases, the prompt space shifts towards an underrepresented distribution in the pre-training corpus, exacerbating attention dilution.
To address this, we introduce a global preference optimization strategy in the second stage.
First, we fine-tune an embedding model via contrastive learning to capture attribute correlations, encouraging the model to prioritize plausible and coherent attribute combinations. This helps steer generation away from implausible combinations rarely seen during pre-training. For example, consider an attribute set that requires the inclusion of an international politics term such as \textit{impose 25\% tariffs}, alongside a medical term like \textit{myocardial infarction}. Such a combination is highly unlikely to appear together in real-world cases.
Second, we promote diversity by selecting the least similar candidate from a pool of generations. This prevents the model from collapsing to a small set of frequent patterns and encourages exploration of less common yet valid combinations.
Together, correlation modeling narrows the search space towards possible regions, while diversity selection expands coverage within that space, enabling the model to retain and balance a large set of attributes during generation.
Our contributions are threefold. First, we design an automated pipeline for dataset construction tailored to extreme constraints, enabling high-quality training and evaluation. Second, we develop a training strategy that integrates reconstruction and RL ~\cite{rafailov2024direct} to address the fundamental challenges in EFCG. Finally, we conduct extensive experiments to validate the proposed framework, providing insights into its efficacy and limitations.


\section{Related Work}
\paragraph{Controllable Text Generation}
CTG tasks involve hard constraints (e.g., text length, keyword inclusion)\cite{takase2019positional, carlsson2022fine} and soft constraints (e.g., sentiment, topic)\cite{gu-etal-2022-distributional, NEURIPS2022_b125999b}. Fine-tuning LLMs with instructional data improves their constraint-following ability~\cite{weller-etal-2020-learning, sanh2021multitask, mishra-etal-2022-cross, DBLP:journals/corr/abs-2402-11905}, but evaluations show LLMs often fail to meet all constraints~\cite{jiang2023followbench, qin2024infobench, ren2025step}. 
Despite this, these works primarily focus on a relatively small number of attributes or conditions, typically from 3 to 5, leaving a gap in understanding LLM's performance under more extreme requirements.

\paragraph{Evaluation of CTG}
Evaluating LLM's adherence to constraints is challenging and typically involves automatic and programmatic assessments using various metrics ~\cite{yao2023collie,zhou2023controlled,chen2022controllable}.~\citet{zhou2023instruction} centers on assessing 25 verifiable instructions.  ~\citet{jiang2023followbench} progressively integrates fine-grained constraints to develop multi-level instructions, thereby enhancing complexity across six distinct types. ~\citet{wen2024benchmarking} constructs a novel benchmark by synthesizing and refining data from the aforementioned benchmarks, with an emphasis on the combinatorial types of constraints. ~\citet{zhang2024cfbench} proposes a comprehensive constraint-following benchmark over 50 NLP tasks. However, none of them investigate the effects of extreme fine-grained attributes.

\paragraph{Multi-objective Alignment}
Recent work~\cite{mudgal2023controlled} focuses on balancing multiple objectives in text generation while maintaining linguistic quality. MORLHF~\cite{zhou2023beyond, rame2024rewarded} optimizes human preferences via reinforcement learning but is costly and unstable. RiC~\cite{yang2024rewards} reduces complexity by using supervised fine-tuning with multi-reward control and dynamic inference adjustment. DeAL ~\cite{huang2024deal} introduces a decoding-time alignment framework for large language models, enabling flexible customization of alignment objectives, such as keyword constraints and abstract goals like harmlessness, without requiring retraining.

\section{Method}
\subsection{Preliminaries}

\begin{figure}[t]
    \centering
        \includegraphics[width=0.49\textwidth]{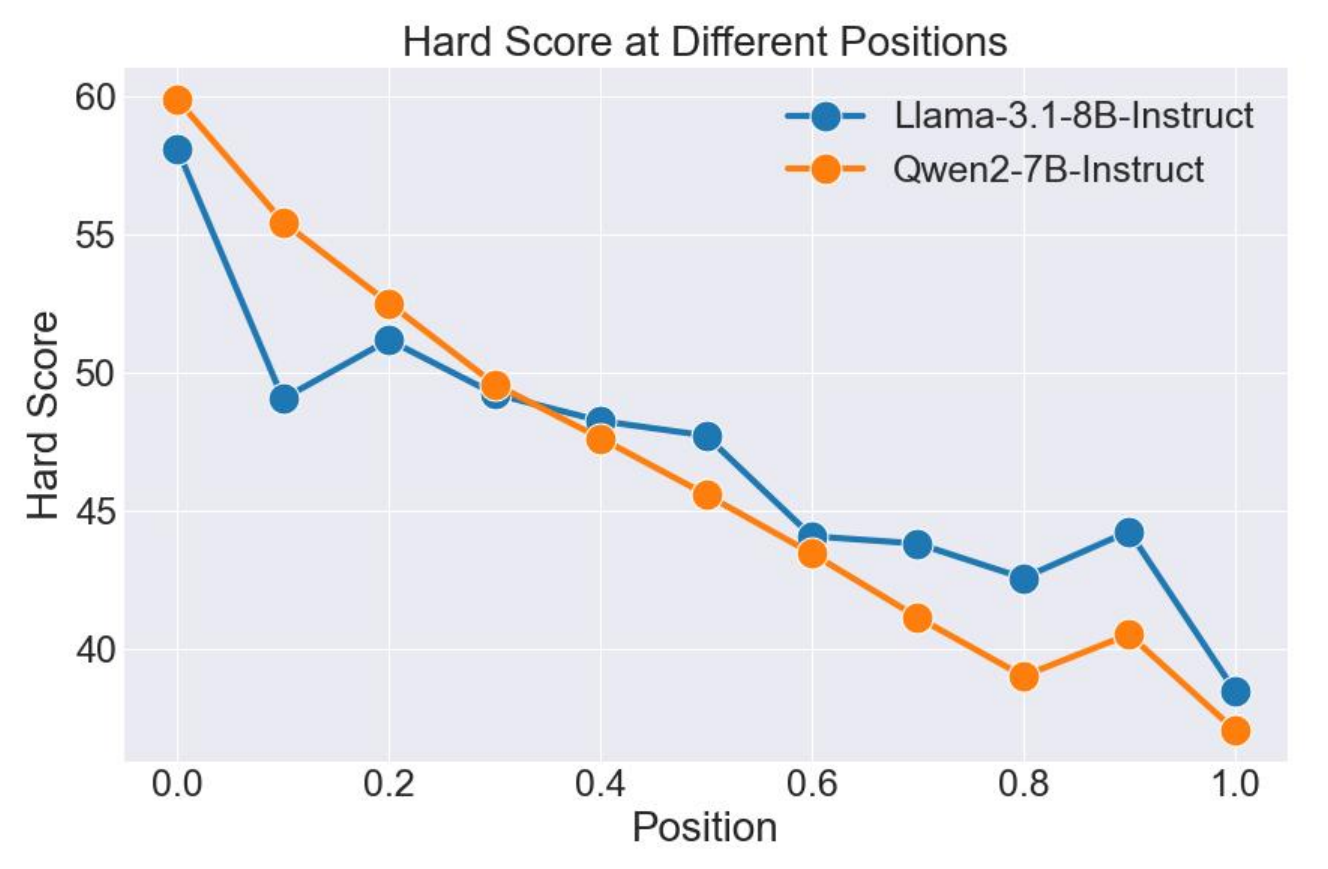}
    \caption{Score degradation as the position of hard attributes shifts in Llama-3.1-8B-Instruct and Qwen2-7B-Instruct, showing a consistent performance drop.}
    \vspace{-0.5em}
    \label{fig:position_bias}
\end{figure}

\paragraph{EFCG: An Overview}
Our extremely EFCG is an extension to controllable text generation CTG, whose goal is to generate an output $Y$ based on a given input $X$ and a set of control conditions $c$. Formally, this can be expressed as:
$$
P(Y \mid X, c)=\prod_{i=1}^n P_\theta\left(Y_i \mid Y_{<i}, X, c\right)
$$

\noindent where $n$ denotes the length of $Y$, and $\theta$ represents the parameters of a language model, and $Y_{<i}$ refers to generated tokens before the $i$-th one. 
In conventional CTG models, $X$ typically serves as a prompt, representing an incomplete text, while $Y$ constitutes its continuation.

Despite advancements made in CTG, existing models often struggle to effectively handle fine-grained control conditions, particularly when the position of attributes within the input context varies. As illustrated in Figure~\ref{fig:position_bias}, this limitation manifests as a significant performance degradation when attributes are positioned away from the beginning of the input context. Such trends highlight that modern language models may not robustly utilize information across the entire input sequence, with biases toward primacy regions of the context window. Addressing this challenge is crucial for improving the robustness and accuracy of EFCG task.

\vspace{-0.5em}
\paragraph{Semantic Similarity}
E5-large~\cite{wang2022text} is a powerful pre-trained encoder optimized for embedding sentences.
In EFCG, we utilize E5-large to encode document-grounded attributes and measure their relationships using cosine similarity. Cosine similarity is a widely used metric to quantify the similarity between two high-dimensional vectors, defined as:
$$
\operatorname{Sim}(\mathbf{u}, \mathbf{v})=\frac{\mathbf{u} \cdot \mathbf{v}}{\|\mathbf{u}\|\|\mathbf{v}\|}
$$

where $\mathbf{u}$ and $\mathbf{v}$ represent the vector embeddings of two attributes. The cosine similarity yields values between -1 (completely dissimilar) and 1 (identical). Using cosine similarity, we identify semantically related attributes and filter out redundant or weakly correlated pairs.

\subsection{UltraGen}
As shown in Figure \ref{fig:pipeline}, our approach, UltraGen, addresses the challenge of EFCG through a two-stage framework.
First, we introduce auto-reconstruction training to align text with a rich set of soft and hard attributes.
Second, we apply global preference optimization to enhance the model’s adaptability to diverse, globally complex attribute compositions. 

\subsubsection{Auto-reconstruction Stage}
The auto-reconstruction stage trains on naturally aligned attribute-text pairs, where attributes are directly extracted from real texts of FineWeb ~\cite{penedo2024the}, a high-quality web corpus, to ensure intrinsic constraint compatibility (i.e., no conflicting attributes exist by construction). Then we train the model to re-generate the text based on the decomposed attributes.
Formally, given a training example $(Y, c) \in D_{\text {UltraBench }}$, the model learns to reconstruct $Y$ by minimizing the negative log-likelihood.

$$
\mathcal{L}_{\mathrm{SFT}}=-\mathbb{E}_{(Y, c)} \log P_\theta(Y \mid c)
$$

This process achieves dual objectives: \textbf{constraint grounding}, which forces the model to internalize the relationships between atomic attributes and their textual realizations, and \textbf{fluency preservation}, which maintains the base model’s generative quality by leveraging the natural language distribution of the original corpus.  The resulting reconstruct model serves as a coherent and constraint-aware initial policy for RL, providing essential prior knowledge for subsequent exploration of complicated constraint combinations.

 \subsubsection{Global Preference Optimization Stage}  
 To extend the foundation capability to global text generation, we first collect a massive pool of attributes from multiple sources.
 The attributes pool integrates diverse sources spanning multiple domains, styles, and formats. Unlike reconstruction, which applies mainly web data, our RL phase leverages data from (1) Books, (2) Academic Papers (arXiv), (3) Social Media (Reddit), (4) Technical Forums (StackExchange), (5) News (CC-News), and (6) Encyclopedic Sources (Wikipedia). This ensures broad coverage of textual variations, enabling the model to generalize across different contexts and constraint types.
 In each iteration, a valid subset of attributes is selected from this pool. Using the auto-reconstruction model, we generate $K$ candidate responses conditioned on the selected attributes. We then apply the CSR metric to identify the preferred and less favorable responses, which are subsequently used for DPO training.

A key challenge in selecting a valid subset of attributes lies in balancing \textbf{topic coherence} and \textbf{anti-redundancy}. Topic coherence requires a high correlation among attributes to ensure interdependent constraints are holistically satisfied. For example, keywords \textit{chain of thought} and \textit{use formal tone} jointly imply technical writing. In contrast, diversity prevents overfitting to frequent patterns and enhances fluency. For example, phrases like \textit{the dreariest place, a dreary day} are redundant and make the text uninformative.
Therefore, our pipeline comprises three key steps:
\vspace{-0.5em}
\paragraph{Attribute Correlation Modeling}
We fine-tune the E5-large encoder using triplet contrastive learning ~\cite{simcse} on document-grounded attributes. For each anchor attribute $\mathcal{A}_i$, a positive pair $\mathcal{A}_j$ shares context from the same document, while a negative pair $\mathcal{A}_k$ is sampled from unrelated contexts. The encoder minimizes the triplet loss, yielding $81.6 \%$ validation accuracy in distinguishing correlated attributes.

\vspace{-0.7em}
\paragraph{Attribute Set Expansion}
The process begins by randomly sampling 2000 seed soft attributes from the attributes pool as the initial attribute set. For each seed attribute $A_i$, we retrieve its top 1024 most correlated candidates using the fine-tuned E5 encoder, where correlation is quantified by the cosine similarity in the correlation representation space using the fine-tuned model. To enforce diversity and minimize semantic redundancy, candidates are iteratively added to the set based on a redundancy score $\operatorname{Sim}\left(A_{\text {candidate }}, A_i\right)$, which is defined as the cosine similarity in their original E5 semantic representation space. Expansion terminates when each set contains a randomly determined number from 10 to 110.

\paragraph{DPO Pair Generation}
For each attribute set, DPO training pairs are constructed by generating $K$ responses using the auto-reconstruction model. The soft and hard attribute scores are obtained by using the Python scripts and GPT-4o (Section \ref{sec:eval}), respectively. The total score is computed by averaging the two scores.

Responses are ranked from highest to lowest based on their scores. The highest-scoring response is chosen, while the lowest-scoring response is rejected. This automated scoring and ranking ensure the selection of thematically coherent and high-quality responses, refining the model’s ability to distinguish and generate optimal outputs.

\subsection{UltraBench}
\subsubsection{Dataset Construction}
\label{sec:dataset_construction}
To support our training framework, we construct two specialized dataset splits named \textbf{UltraBench}, derived from FineWeb \cite{penedo2024the} and multiple sources.
The UltraBench dataset is designed to evaluate and train models on extremely fine-grained controllable text generation. Its construction involves Two stages, as detailed below.


\paragraph{Attribute Extraction}

Attributes were categorized into two types:
\begin{enumerate}
    \item \textbf{Soft attributes:} (e.g., style, tone, content) were inferred using GPT-4o~\cite{achiam2023gpt} to capture semantic properties. For example, a soft attribute might describe a passage as a \textit{vivid personal narrative focused on childbirth experience}.
    \item \textbf{Hard attributes:} consist of programmatically verifiable constraints extracted directly from the text. These included keyword requirements (e.g., \textit{include sustainability}), structural rules (e.g., \textit{generate exactly three paragraphs}), and syntactic directives (e.g., \textit{use all lowercase letters}).
\end{enumerate}
For the FineWeb split, we use each attribute set along with its corresponding raw text to perform the auto-reconstruction stage. For the Multi-sources split, we aggregate and de-duplicate all decomposed attributes to form a global attribute pool.

\paragraph{Consistency Verification}
To ensure the reliability of soft attribute extraction, we conducted a human evaluation on a randomly selected subset of 100 documents. Human experts assessed whether the extracted attributes accurately reflected the underlying text. We computed the Agreement Rate (AR), defined as the proportion of samples where automated extractions matched the original raw text. This process achieved an AR of 96.5\%, indicating a strong alignment between attributes and original text.

\subsubsection{Dataset Statistics}
\paragraph{Overall Statistics}
In Appendix ~\ref{appendix:statistics}, we summarize the dataset details. Table \ref{tab:overall_statistic} details the composition of UltraBench, with separate configurations for reconstruction and multi-sources subsets. Table \ref{tab:rl_domains} further analyzes the multi-sources subset’s domain distribution, while Table \ref{tab:quality_metrics} quantifies quality control metrics. 

\paragraph{Compared with Other Benchmarks}

\begin{figure}[t] 
    \centering
        \includegraphics[width=0.49\textwidth]{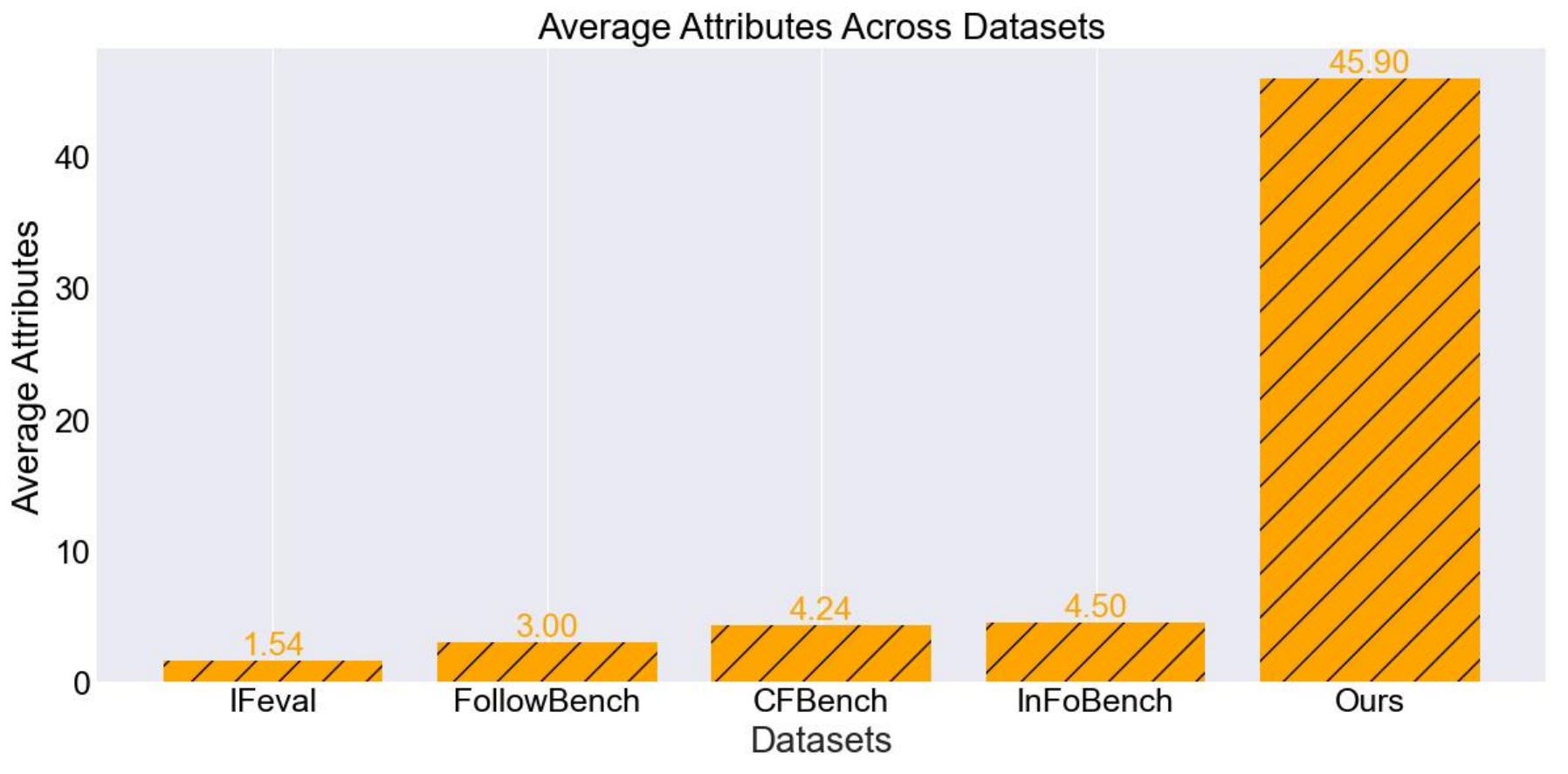}
    \caption{Comparison of average attributes across datasets.}
    \vspace{-1em}
    \label{fig:intro}
\end{figure}
Table ~\ref{tab:comparison} provides a detailed comparison of our dataset with other relevant works. 
While IFeval and FollowBench include synthesized data (Synt.), they fall short in capturing the diversity and complexity required for evaluating real-world applications.
Another key strength of our dataset lies in the average number of attributes per sample, where we achieve a remarkable value of 45.9 and 29.9 on two splits, far exceeding the benchmarks' maximum of 4.5. This demonstrates the ability of our dataset for evaluating tasks requiring fine-grained attribute understanding.

\subsubsection{Evaluation Protocol}
\label{sec:eval}
To rigorously evaluate EFCG capabilities, we use two evaluation metrics:

\paragraph{Constraint Satisfaction Rate (CSR)}

For a given instruction with both soft and hard constraints, we compute the CSR as follows:

$$
\mathrm{CSR}=\frac{1}{m} \sum_{i=1}^m \frac{1}{n^{(i)}} \sum_{j=1}^{n^{(i)}} s_j^{(i)}
$$

where $s_j^{(i)}=1$ if the $j$-th constraint for the $i$-th instruction is satisfied, and 0 otherwise. Here, $n^{(i)}$ is the number of constraints (hard or soft) for instruction $i$, and $m$ is the total number of evaluated instructions.
\begin{enumerate}
    \item \textbf{Hard Constraint Verification}: For programmatically verifiable constraints, we perform deterministic checks via Python scripts. Due to the significant imbalance in hard attributes, we adopt macro accuracy to ensure fair evaluation. Macro accuracy computes the average CSR across different types, giving equal weight to each type regardless of its frequency.
    
    \item \textbf{Soft Constraint Evaluation}: For semantic constraints, we employ an LLM-based judge (GPT-4o), assigning a binary score (0 or 1) to each constraint. 
    We validate the quality of the LLM-based judges on a randomly selected set of 100 samples. By calculating the Cohen’s Kappa coefficient between the scores of LLM-based judge and human experts, we found a strong agreement (84.55\%) between the automatic evaluation and human experts' assessment. 

\end{enumerate}

\paragraph{BERTScore} 
In the auto-reconstruction phase, we also use BERTScore~\cite{bertscore} to measure the quality of the reconstructed text. BERTScore leverages the contextual embeddings from pre-trained language models to capture semantic similarity. BERTScore is widely used in text generation tasks, as it aligns better with human judgments of semantic quality compared to traditional n-gram overlap-based metrics.

\section{Experiments}
\subsection{Experiment Setup} 
\paragraph{Models.}

\begin{table*}[htbp]
\newcolumntype{g}{>{\columncolor{green!10}}c}
\newcolumntype{b}{>{\columncolor{blue!10}}c}
\renewcommand{\arraystretch}{1.22} 
\small
\resizebox{\textwidth}{!}
{
\begin{tabular}{llccccccc}

\toprule
&  \multirow{2}*{Model} & \multicolumn{4}{c}{\textbf{FineWeb Split}} & \multicolumn{3}{c}{\textbf{Multi-source Split}} \\
\cmidrule(l){3-6} \cmidrule(l){7-9} 
& & Overall Score & Soft Score & Hard Score & BERTScore F1    & Overall Score & Soft Score & Hard Score   \\ 
 
 \midrule

 \multirow{3}*{\rotatebox{90}{Main}}& Base Model  & 50.30 & 67.08 & 33.51 & 59.92  & 37.45       & 36.10            & 38.79            \\

& UltraGen (AR)  & 56.05 & 81.44  & 30.65 & 62.00    & 50.15         & 62.41               & 37.89            \\
& UltraGen (AR+GPO) &  59.61 & 84.33 & 34.89 & 61.22    &  57.23       & 69.01               & 45.44            \\ 
\midrule
\multirow{4}*{\rotatebox{90}{Ablation}} & AR (Few Constraints) & 48.25 & 74.09 & 22.41 & 60.10    & 38.38         &  46.00        & 30.76             \\
& GPO & 55.57 & 74.50 & 36.63 & 60.59 & 42.44 & 51.00 & 33.86 \\
& AR+GPO (Random Sampling) &  59.77 & 85.42 & 34.11 & 60.56 & 55.24         & 68.01            & 42.47            \\ 
& AR+GPO (High Similarity) &  59.44 & 83.22 & 35.65 & 60.85 & 55.45        & 66.05               & 44.85             \\ 
& AR+GPO (Low Correlation) &  58.91 & 83.59 & 34.23 & 60.00 & 54.47         & 65.22               & 43.71             \\ 

\bottomrule

\end{tabular}
}
\caption{Performance scores for Llama-3.2-3B-Instruct models on the validation set under different evaluation conditions across FineWeb and Global splits.}
\label{tab:ultrabench}
\end{table*}

Our experiments evaluate the EFCG task using one mainstream instruction-tuned base model: Llama-3.2-3B-Instruct ~\cite{dubey2024llama}, chosen for its demonstrated proficiency in instruction-following tasks within the 3B parameter range. To systematically assess the impact of our methodology, we compare three training paradigms: (1) \textbf{BASE}, which directly employs the unmodified base models to establish a performance baseline; (2) \textbf{AR}, where models undergo the auto-reconstruction stage on our meticulously constructed FineWeb dataset (§3.2), enriched with fine-grained attributes to enhance multi-constraint adherence; and (3) \textbf{AR+GPO}, a hybrid optimization approach combining direct preference optimization with global embedding space adaption.

\subsection{Evaluation Results on UltraBench}

Our experimental findings, summarized in Table \ref{tab:ultrabench}, demonstrate the substantial advancements achieved by applying the UltraGen paradigm to EFCG. The evaluation leverages the validation set of FineWeb and Global splits to assess model performance under both local and global constraints.

The application of AR yielded significant improvements over the base model. On the FineWeb split, the AR model attained an overall score of 56.05, representing a relative improvement of 11.4\%. The soft score rose to 81.44, indicating enhanced adherence to semantic and stylistic attributes, while the hard score increased to 30.65, reflecting better performance on programmatically verifiable constraints. On the Global split, the AR model demonstrated its ability to generalize, achieving an overall score of 50.15.

Further optimization through GPO demonstrated remarkable performance on the Global split, where the model achieved an overall score of 57.23 and an impressive hard score of 45.44. This highlights the model's robust generalization and optimization capabilities when dealing with diverse and challenging global constraints. Notably, despite being trained on the Global split, the AR+GPO model exhibited strong performance on the FineWeb split as well, achieving an overall score of 59.61, a soft score of 84.33, and a hard score of 34.89. This result underscores the model's ability to transfer its learned capabilities from the broader and more diverse Global split to the more localized FineWeb split.

\paragraph{Ablation}
To evaluate the contribution of key components in our UltraGen framework, we conducted ablation studies by systematically modifying the training process. We tested the impact of reducing the number of attributes during AR, removing the AR stage, replacing curated attributes with random sampling, and eliminating the high-correlation or low-redundancy selection steps. The results demonstrate that both AR and GPO stages are crucial for achieving strong performance, as reducing constraints, removing correlation modeling, or neglecting redundancy minimization leads to performance degradation.
\subsection{Data Synthesis Improvement}

\begin{table}[htbp]
\centering
\small
\resizebox{0.48\textwidth}{!}{
\begin{tabular}{lccc}
\toprule
\textbf{Dataset (Domain)} & \textbf{Base} &  \textbf{AR} & \textbf{AR+GPO} \\ \midrule
Emotion (Tweet Emotion) & 28.25 & \textbf{42.30}  & 38.65 \\
Hillary (Tweet Stance)  & 55.93  & 45.76 & \textbf{58.31} \\
AG-News (News Topic) & 80.03 & 79.96 &\textbf{83.28} \\
TREC (Question Type) & 38.00  & 51.20  & \textbf{51.40} \\ 
\midrule
Average   & 50.55 & 54.81 & \textbf{57.91} \\
\bottomrule
\end{tabular}
}
\caption{Performance comparison for data synthesis.}
\vspace{-1em}
\label{tab:data_synthesis}
\end{table}

To demonstrate the improvement in the usage of texts synthesized by UltraGen, we utilize several diverse well-established text classification benchmarks to test the data synthesis capability, such as sentiment analysis \textbf{(1) Emotion} ~\cite{saravia-etal-2018-carer}, attitude classification towards a particular public figure \textbf{(2) Hillary} ~\cite{barbieri2020tweeteval}, topic classification \textbf{(3) AG News} ~\cite{Zhang2015CharacterlevelCN}, question type classification \textbf{(4) TREC} ~\cite{li-roth-2002-learning}.

For each dataset, we analyze the unique properties and paraphrase these properties as hard and soft attributes. Then using a uniform prompt tailored for each dataset, we generate 2,000 synthetic samples per dataset. These generated samples are then used to train a classifier, which is subsequently evaluated on the original test set of the dataset. This procedure allows for a fair comparison of model performance on synthetic data. 

The results, summarized in Table \ref{tab:data_synthesis}, demonstrate the superior generalization ability of the AR+GPO model trained on the Global split. Notably, the AR+GPO model achieved the highest average score of 57.91 across the benchmarks, significantly outperforming both the base model and the AR models. While the AR model’s performance stagnated (45.76, lower than the original one) on the Hillary benchmark, reflecting a focus on localized attributes, the AR+GPO model excelled with a score of 58.31, indicating its generalization and adaptability beyond localized training objectives.

\subsection{Trade-Offs in EFCG}
\begin{figure}[t]
    \centering
        \includegraphics[width=0.49\textwidth]{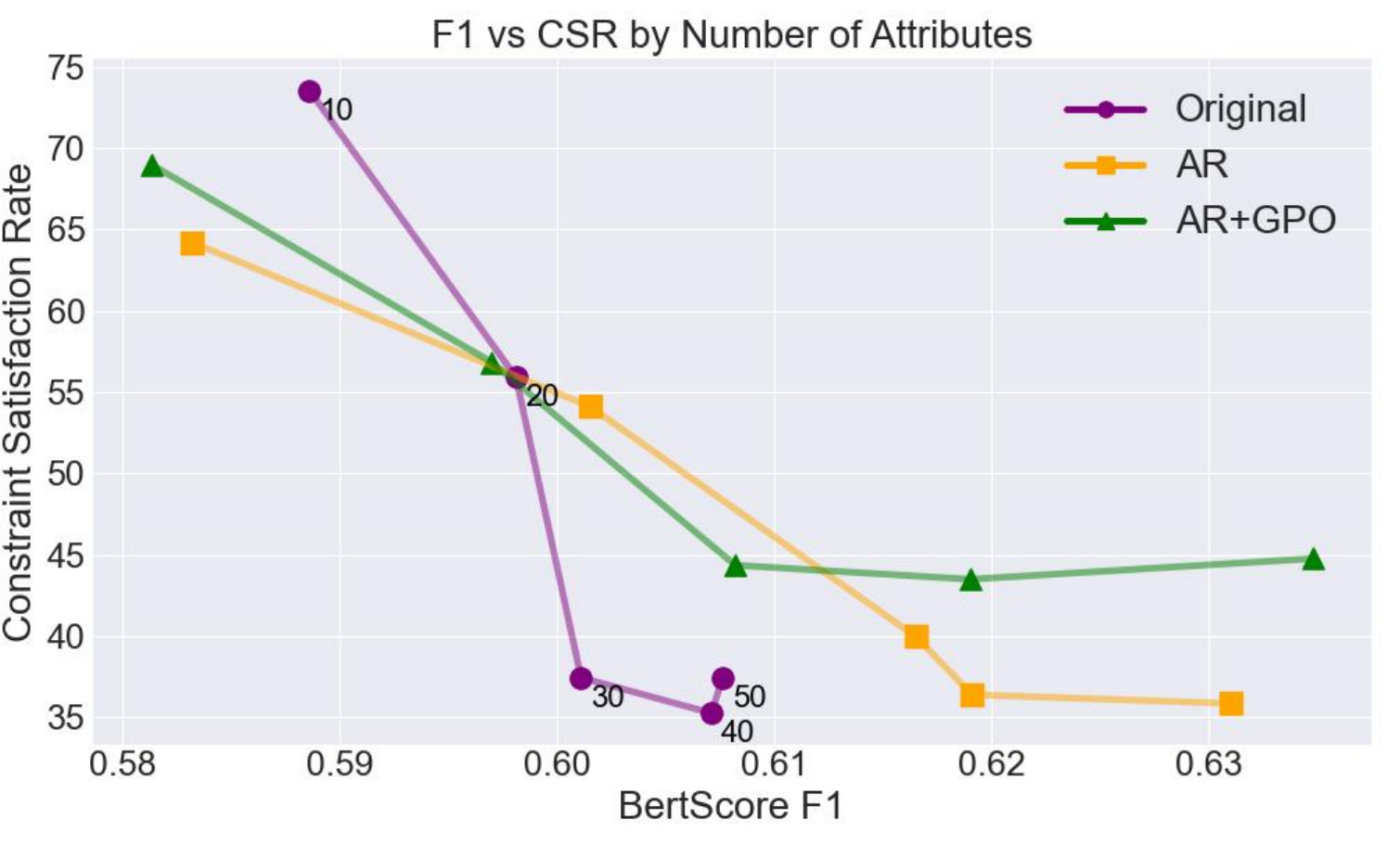}
    \caption{The Trade-off between F1 score and CSR. While BERTScore tends to improve with more attributes, CSR declines}
    \vspace{-1.5em}
    \label{fig:tradeoff}
\end{figure}

Figure~\ref{fig:tradeoff} illustrates the interplay between BERTScore and CSR across different numbers of attributes from 10 to 50 for each model. As the figure shows, increasing the number of attributes presents a clear double-edged effect: while more attributes can enhance fine-grained control (e.g., higher F1 score) over the generated text, the added complexity makes it more difficult for the model to maintain high constraint adherence.

\paragraph{Better Multi-Objective Alignment Under EFCG.}
\begin{figure*}[htbp]
    \centering
        \includegraphics[width=0.98\textwidth]{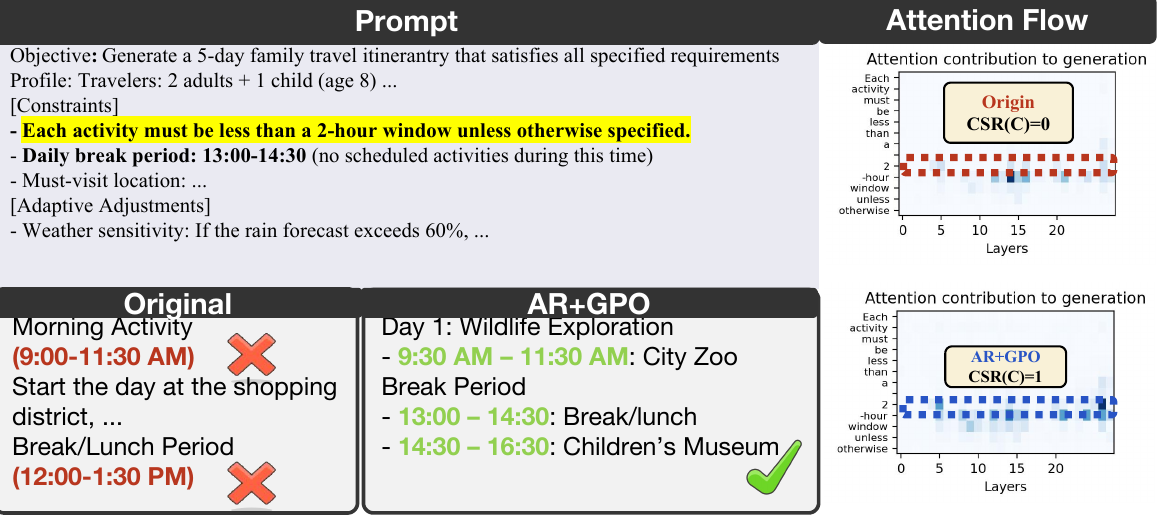}
    \caption{In a case study on travel itinerary generation, the attention flow illustrates improved constraint awareness in AR+GPO.}
    \vspace{-1em}
    \label{fig:case_study}
\end{figure*}

When looking at the 30, 40, and 50 attribute conditions:
AR+GPO consistently attains CSR values 5--10 points higher than the other two models without sacrificing F1.
For example, at 50 attributes, AR+GPO’s CSR (44.76\%) is considerably above AR’s (35.86\%) and Original’s (37.40\%), while also delivering the highest F1 (0.6348 vs. 0.6310 for AR and 0.6076 for Original).

This pattern illustrates a more favorable trade-off for AR+GPO: it does not simply chase high BERTScore by ignoring constraints, nor does it force all constraints at the expense of overall text quality. Instead, AR+GPO’s global optimization helps coordinate multiple constraints while retaining strong semantic alignment. In contrast, AR appears effective at moderate attribute counts but loses ground on CSR once the load goes beyond 30 attributes, and the Original model experiences an even steeper decline.

\section{Analysis}

\subsection{UltraBench Mitigates Performance Degradation Across Different Positions}
\begin{figure}[htbp]
    \centering
        \includegraphics[width=0.49\textwidth]{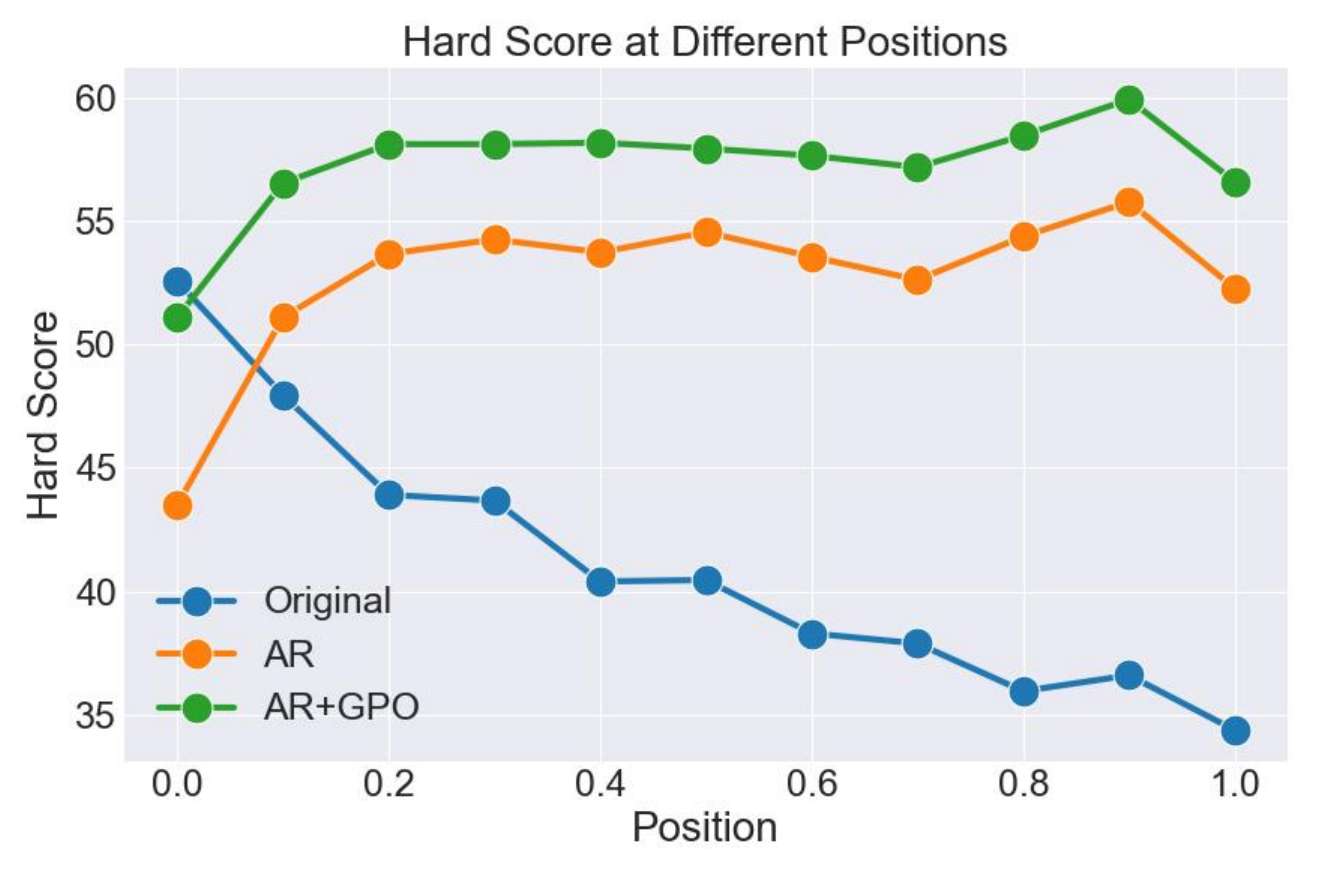}
    \caption{Hard score across different positions, showing that our approach (AR+GPO) effectively mitigates performance degradation.}
    \vspace{-1em}
    \label{fig:lost_in_the_middle}
\end{figure}
Our method effectively mitigates the sensitivity to positional changes in hard attributes. As shown in Figure ~\ref{fig:position_bias}, baseline models such as Llama-3.1-8B-Instruct and Qwen2-7B-Instruct exhibit a significant drop in hard scores as the position increases, indicating a degradation in performance when hard attributes appear later in the input. In contrast, our approach significantly stabilizes performance across all positions (Figure ~\ref{fig:lost_in_the_middle}). The introduction of AR already improves robustness compared to the original model, and the addition of GPO further enhances consistency, maintaining high hard scores even at later positions. This demonstrates that our approach effectively addresses the position sensitivity issue, ensuring more reliable model performance regardless of attribute placement.
\subsection{A Real-world Travel Case}
\label{sec:case_study}
To further evaluate our approach, we analyze a real-world travel planning scenario where the itinerary must satisfy over thirty attributes. One crucial constraint is that \textit{each activity should be less than two hours long}. The authors examined the response generated by three models. We observe that only the AR+GPO model consistently generates activities that adhere to this constraint, whereas the original model and AR model occasionally violate it. To gain deeper insights, we provide the user prompt along with a partially generated response (e.g., "14:00 PM - 1") and examine the attention flow distribution at this intermediate step. As illustrated in Figure \ref{fig:case_study}, the AR+GPO model exhibits significantly higher attention weights on constraint-related tokens (e.g., "2"), suggesting that it effectively retains and incorporates constraint-relevant information during generation. In contrast, the original model's attention weights are relatively weak, indicating a lower degree of constraint awareness.

\section{Conclusion}
We proposed UltraGen, a two-stage framework for extremely fine-grained controllable generation. The Auto-Reconstruction stage trains LLMs to align with both soft and hard attributes, while Global Preference Optimization further enhances constraint satisfaction under diverse attribute combinations. Experiments on UltraBench demonstrate that UltraGen significantly improves both constraint adherence and text quality. 

\section*{Limitations}

While UltraGen demonstrates strong performance in handling extremely fine-grained controllable generation, several limitations remain. First, the set of hard attributes used in this work, though diverse and practical, primarily focuses on structural and keyword constraints; future work could explore more complex and domain-specific hard constraints to further stress-test model capabilities. Second, although our attribute correlation and diversity strategies reduce implausible combinations, ensuring absolute coherence across a large number of constraints remains an open challenge.

\bibliography{custom}

\begin{thebibliography}{36}
\providecommand{\natexlab}[1]{#1}

\bibitem[{Achiam et~al.(2023)Achiam, Adler, Agarwal, Ahmad, Akkaya, Aleman, Almeida, Altenschmidt, Altman, Anadkat et~al.}]{achiam2023gpt}
Josh Achiam, Steven Adler, Sandhini Agarwal, Lama Ahmad, Ilge Akkaya, Florencia~Leoni Aleman, Diogo Almeida, Janko Altenschmidt, Sam Altman, Shyamal Anadkat, et~al. 2023.
\newblock Gpt-4 technical report.
\newblock \emph{arXiv preprint arXiv:2303.08774}.

\bibitem[{Barbieri et~al.(2020)Barbieri, Camacho-Collados, Neves, and Espinosa-Anke}]{barbieri2020tweeteval}
Francesco Barbieri, Jose Camacho-Collados, Leonardo Neves, and Luis Espinosa-Anke. 2020.
\newblock Tweeteval: Unified benchmark and comparative evaluation for tweet classification.
\newblock \emph{arXiv preprint arXiv:2010.12421}.

\bibitem[{Carlsson et~al.(2022)Carlsson, {\"O}hman, Liu, Verlinden, Nivre, and Sahlgren}]{carlsson2022fine}
Fredrik Carlsson, Joey {\"O}hman, Fangyu Liu, Severine Verlinden, Joakim Nivre, and Magnus Sahlgren. 2022.
\newblock Fine-grained controllable text generation using non-residual prompting.
\newblock In \emph{Proceedings of the 60th Annual Meeting of the Association for Computational Linguistics (Volume 1: Long Papers)}, pages 6837--6857.

\bibitem[{Chen et~al.(2022)Chen, Li, Chen, and Narasimhan}]{chen2022controllable}
Howard Chen, Huihan Li, Danqi Chen, and Karthik Narasimhan. 2022.
\newblock Controllable text generation with language constraints.
\newblock \emph{arXiv preprint arXiv:2212.10466}.

\bibitem[{Dubey et~al.(2024)Dubey, Jauhri, Pandey, Kadian, Al-Dahle, Letman, Mathur, Schelten, Yang, Fan et~al.}]{dubey2024llama}
Abhimanyu Dubey, Abhinav Jauhri, Abhinav Pandey, Abhishek Kadian, Ahmad Al-Dahle, Aiesha Letman, Akhil Mathur, Alan Schelten, Amy Yang, Angela Fan, et~al. 2024.
\newblock The llama 3 herd of models.
\newblock \emph{arXiv preprint arXiv:2407.21783}.

\bibitem[{Gao et~al.(2021)Gao, Yao, and Chen}]{simcse}
Tianyu Gao, Xingcheng Yao, and Danqi Chen. 2021.
\newblock \href {https://doi.org/10.18653/V1/2021.EMNLP-MAIN.552} {Simcse: Simple contrastive learning of sentence embeddings}.
\newblock In \emph{Proceedings of the 2021 Conference on Empirical Methods in Natural Language Processing, {EMNLP} 2021, Virtual Event / Punta Cana, Dominican Republic, 7-11 November, 2021}, pages 6894--6910. Association for Computational Linguistics.

\bibitem[{Gu et~al.(2022)Gu, Feng, Ma, Zhang, Gong, and Qin}]{gu-etal-2022-distributional}
Yuxuan Gu, Xiaocheng Feng, Sicheng Ma, Lingyuan Zhang, Heng Gong, and Bing Qin. 2022.
\newblock \href {https://aclanthology.org/2022.emnlp-main.67} {A distributional lens for multi-aspect controllable text generation}.
\newblock In \emph{Proceedings of the 2022 Conference on Empirical Methods in Natural Language Processing}, pages 1023--1043, Abu Dhabi, United Arab Emirates. Association for Computational Linguistics.

\bibitem[{Huang et~al.(2024)Huang, Sengupta, Bonadiman, Lai, Gupta, Pappas, Mansour, Kirchhoff, and Roth}]{huang2024deal}
James~Y Huang, Sailik Sengupta, Daniele Bonadiman, Yi-an Lai, Arshit Gupta, Nikolaos Pappas, Saab Mansour, Katrin Kirchhoff, and Dan Roth. 2024.
\newblock Deal: Decoding-time alignment for large language models.
\newblock \emph{arXiv preprint arXiv:2402.06147}.

\bibitem[{Jiang et~al.(2024)Jiang, Wang, Wu, Zhong, Zeng, Gao, Li, Jiang, Shang, Tang, Liu, and Wang}]{DBLP:journals/corr/abs-2402-11905}
Yuxin Jiang, Yufei Wang, Chuhan Wu, Wanjun Zhong, Xingshan Zeng, Jiahui Gao, Liangyou Li, Xin Jiang, Lifeng Shang, Ruiming Tang, Qun Liu, and Wei Wang. 2024.
\newblock \href {https://doi.org/10.48550/ARXIV.2402.11905} {Learning to edit: Aligning llms with knowledge editing}.
\newblock \emph{CoRR}, abs/2402.11905.

\bibitem[{Jiang et~al.(2023)Jiang, Wang, Zeng, Zhong, Li, Mi, Shang, Jiang, Liu, and Wang}]{jiang2023followbench}
Yuxin Jiang, Yufei Wang, Xingshan Zeng, Wanjun Zhong, Liangyou Li, Fei Mi, Lifeng Shang, Xin Jiang, Qun Liu, and Wei Wang. 2023.
\newblock Followbench: A multi-level fine-grained constraints following benchmark for large language models.
\newblock \emph{arXiv preprint arXiv:2310.20410}.

\bibitem[{Li and Roth(2002)}]{li-roth-2002-learning}
Xin Li and Dan Roth. 2002.
\newblock \href {https://www.aclweb.org/anthology/C02-1150} {Learning question classifiers}.
\newblock In \emph{{COLING} 2002: The 19th International Conference on Computational Linguistics}.

\bibitem[{Liu et~al.(2024)Liu, Lin, Hewitt, Paranjape, Bevilacqua, Petroni, and Liang}]{lost-in-the-middle}
Nelson~F Liu, Kevin Lin, John Hewitt, Ashwin Paranjape, Michele Bevilacqua, Fabio Petroni, and Percy Liang. 2024.
\newblock Lost in the middle: How language models use long contexts.
\newblock \emph{Transactions of the Association for Computational Linguistics}, 12:157--173.

\bibitem[{Lu et~al.(2022)Lu, Welleck, Hessel, Jiang, Qin, West, Ammanabrolu, and Choi}]{NEURIPS2022_b125999b}
Ximing Lu, Sean Welleck, Jack Hessel, Liwei Jiang, Lianhui Qin, Peter West, Prithviraj Ammanabrolu, and Yejin Choi. 2022.
\newblock \href {https://proceedings.neurips.cc/paper_files/paper/2022/file/b125999bde7e80910cbdbd323087df8f-Paper-Conference.pdf} {Quark: Controllable text generation with reinforced unlearning}.
\newblock In \emph{Advances in Neural Information Processing Systems}, volume~35, pages 27591--27609. Curran Associates, Inc.

\bibitem[{Mishra et~al.(2022)Mishra, Khashabi, Baral, and Hajishirzi}]{mishra-etal-2022-cross}
Swaroop Mishra, Daniel Khashabi, Chitta Baral, and Hannaneh Hajishirzi. 2022.
\newblock \href {https://doi.org/10.18653/v1/2022.acl-long.244} {Cross-task generalization via natural language crowdsourcing instructions}.
\newblock In \emph{Proceedings of the 60th Annual Meeting of the Association for Computational Linguistics (Volume 1: Long Papers)}, pages 3470--3487, Dublin, Ireland. Association for Computational Linguistics.

\bibitem[{Mudgal et~al.(2023)Mudgal, Lee, Ganapathy, Li, Wang, Huang, Chen, Cheng, Collins, Strohman et~al.}]{mudgal2023controlled}
Sidharth Mudgal, Jong Lee, Harish Ganapathy, YaGuang Li, Tao Wang, Yanping Huang, Zhifeng Chen, Heng-Tze Cheng, Michael Collins, Trevor Strohman, et~al. 2023.
\newblock Controlled decoding from language models.
\newblock \emph{arXiv preprint arXiv:2310.17022}.

\bibitem[{Penedo et~al.(2024)Penedo, Kydl{\'\i}{\v{c}}ek, allal, Lozhkov, Mitchell, Raffel, Werra, and Wolf}]{penedo2024the}
Guilherme Penedo, Hynek Kydl{\'\i}{\v{c}}ek, Loubna~Ben allal, Anton Lozhkov, Margaret Mitchell, Colin Raffel, Leandro~Von Werra, and Thomas Wolf. 2024.
\newblock \href {https://openreview.net/forum?id=n6SCkn2QaG} {The fineweb datasets: Decanting the web for the finest text data at scale}.
\newblock In \emph{The Thirty-eight Conference on Neural Information Processing Systems Datasets and Benchmarks Track}.

\bibitem[{Qin et~al.(2024)Qin, Song, Hu, Yao, Cho, Wang, Wu, Liu, Liu, and Yu}]{qin2024infobench}
Yiwei Qin, Kaiqiang Song, Yebowen Hu, Wenlin Yao, Sangwoo Cho, Xiaoyang Wang, Xuansheng Wu, Fei Liu, Pengfei Liu, and Dong Yu. 2024.
\newblock Infobench: Evaluating instruction following ability in large language models.
\newblock \emph{arXiv preprint arXiv:2401.03601}.

\bibitem[{Rafailov et~al.(2024)Rafailov, Sharma, Mitchell, Manning, Ermon, and Finn}]{rafailov2024direct}
Rafael Rafailov, Archit Sharma, Eric Mitchell, Christopher~D Manning, Stefano Ermon, and Chelsea Finn. 2024.
\newblock Direct preference optimization: Your language model is secretly a reward model.
\newblock \emph{Advances in Neural Information Processing Systems}, 36.

\bibitem[{Rame et~al.(2024)Rame, Couairon, Dancette, Gaya, Shukor, Soulier, and Cord}]{rame2024rewarded}
Alexandre Rame, Guillaume Couairon, Corentin Dancette, Jean-Baptiste Gaya, Mustafa Shukor, Laure Soulier, and Matthieu Cord. 2024.
\newblock Rewarded soups: towards pareto-optimal alignment by interpolating weights fine-tuned on diverse rewards.
\newblock \emph{Advances in Neural Information Processing Systems}, 36.

\bibitem[{Ren et~al.(2025)Ren, Zeng, He, Liang, Xiao, Zhou, Sun, and Yu}]{ren2025step}
Qingyu Ren, Jie Zeng, Qianyu He, Jiaqing Liang, Yanghua Xiao, Weikang Zhou, Zeye Sun, and Fei Yu. 2025.
\newblock Step-by-step mastery: Enhancing soft constraint following ability of large language models.
\newblock \emph{arXiv preprint arXiv:2501.04945}.

\bibitem[{Sanh et~al.(2021)Sanh, Webson, Raffel, Bach, Sutawika, Alyafeai, Chaffin, Stiegler, Scao, Raja et~al.}]{sanh2021multitask}
Victor Sanh, Albert Webson, Colin Raffel, Stephen~H Bach, Lintang Sutawika, Zaid Alyafeai, Antoine Chaffin, Arnaud Stiegler, Teven~Le Scao, Arun Raja, et~al. 2021.
\newblock Multitask prompted training enables zero-shot task generalization.
\newblock \emph{arXiv preprint arXiv:2110.08207}.

\bibitem[{Saravia et~al.(2018)Saravia, Liu, Huang, Wu, and Chen}]{saravia-etal-2018-carer}
Elvis Saravia, Hsien-Chi~Toby Liu, Yen-Hao Huang, Junlin Wu, and Yi-Shin Chen. 2018.
\newblock \href {https://doi.org/10.18653/v1/D18-1404} {{CARER}: Contextualized affect representations for emotion recognition}.
\newblock In \emph{Proceedings of the 2018 Conference on Empirical Methods in Natural Language Processing}, pages 3687--3697, Brussels, Belgium. Association for Computational Linguistics.

\bibitem[{Song et~al.(2024)Song, Yu, Li, Yu, Huang, Li, and Wang}]{song2024preference}
Feifan Song, Bowen Yu, Minghao Li, Haiyang Yu, Fei Huang, Yongbin Li, and Houfeng Wang. 2024.
\newblock Preference ranking optimization for human alignment.
\newblock In \emph{Proceedings of the AAAI Conference on Artificial Intelligence}, volume~38, pages 18990--18998.

\bibitem[{Takase and Okazaki(2019)}]{takase2019positional}
Sho Takase and Naoaki Okazaki. 2019.
\newblock Positional encoding to control output sequence length.
\newblock In \emph{Proceedings of NAACL-HLT}, pages 3999--4004.

\bibitem[{Wang et~al.(2022)Wang, Yang, Huang, Jiao, Yang, Jiang, Majumder, and Wei}]{wang2022text}
Liang Wang, Nan Yang, Xiaolong Huang, Binxing Jiao, Linjun Yang, Daxin Jiang, Rangan Majumder, and Furu Wei. 2022.
\newblock Text embeddings by weakly-supervised contrastive pre-training.
\newblock \emph{arXiv preprint arXiv:2212.03533}.

\bibitem[{Wang et~al.(2023)Wang, Zhong, Li, Mi, Zeng, Huang, Shang, Jiang, and Liu}]{wang2023aligning}
Yufei Wang, Wanjun Zhong, Liangyou Li, Fei Mi, Xingshan Zeng, Wenyong Huang, Lifeng Shang, Xin Jiang, and Qun Liu. 2023.
\newblock Aligning large language models with human: A survey.
\newblock \emph{arXiv preprint arXiv:2307.12966}.

\bibitem[{Weller et~al.(2020)Weller, Lourie, Gardner, and Peters}]{weller-etal-2020-learning}
Orion Weller, Nicholas Lourie, Matt Gardner, and Matthew~E. Peters. 2020.
\newblock \href {https://doi.org/10.18653/v1/2020.emnlp-main.105} {Learning from task descriptions}.
\newblock In \emph{Proceedings of the 2020 Conference on Empirical Methods in Natural Language Processing (EMNLP)}, pages 1361--1375, Online. Association for Computational Linguistics.

\bibitem[{Wen et~al.(2024)Wen, Ke, Gu, Wu, Huang, Zhou, Li, Hu, Gao, Xu et~al.}]{wen2024benchmarking}
Bosi Wen, Pei Ke, Xiaotao Gu, Lindong Wu, Hao Huang, Jinfeng Zhou, Wenchuang Li, Binxin Hu, Wendy Gao, Jiaxin Xu, et~al. 2024.
\newblock Benchmarking complex instruction-following with multiple constraints composition.
\newblock \emph{arXiv preprint arXiv:2407.03978}.

\bibitem[{Yang et~al.(2024)Yang, Pan, Luo, Qiu, Zhong, Yu, and Chen}]{yang2024rewards}
Rui Yang, Xiaoman Pan, Feng Luo, Shuang Qiu, Han Zhong, Dong Yu, and Jianshu Chen. 2024.
\newblock Rewards-in-context: Multi-objective alignment of foundation models with dynamic preference adjustment.
\newblock \emph{arXiv preprint arXiv:2402.10207}.

\bibitem[{Yao et~al.(2023)Yao, Chen, Hanjie, Yang, and Narasimhan}]{yao2023collie}
Shunyu Yao, Howard Chen, Austin~W Hanjie, Runzhe Yang, and Karthik Narasimhan. 2023.
\newblock Collie: Systematic construction of constrained text generation tasks.
\newblock \emph{arXiv preprint arXiv:2307.08689}.

\bibitem[{Zhang et~al.(2024)Zhang, Shen, Luo, Zhang, Liang, Yang, Lin, Qiao, Chen, Cui et~al.}]{zhang2024cfbench}
Tao Zhang, Yanjun Shen, Wenjing Luo, Yan Zhang, Hao Liang, Fan Yang, Mingan Lin, Yujing Qiao, Weipeng Chen, Bin Cui, et~al. 2024.
\newblock Cfbench: A comprehensive constraints-following benchmark for llms.
\newblock \emph{arXiv preprint arXiv:2408.01122}.

\bibitem[{Zhang et~al.(2020)Zhang, Kishore, Wu, Weinberger, and Artzi}]{bertscore}
Tianyi Zhang, Varsha Kishore, Felix Wu, Kilian~Q. Weinberger, and Yoav Artzi. 2020.
\newblock \href {https://openreview.net/forum?id=SkeHuCVFDr} {Bertscore: Evaluating text generation with {BERT}}.
\newblock In \emph{8th International Conference on Learning Representations, {ICLR} 2020, Addis Ababa, Ethiopia, April 26-30, 2020}. OpenReview.net.

\bibitem[{Zhang et~al.(2015)Zhang, Zhao, and LeCun}]{Zhang2015CharacterlevelCN}
Xiang Zhang, Junbo~Jake Zhao, and Yann LeCun. 2015.
\newblock Character-level convolutional networks for text classification.
\newblock In \emph{NIPS}.

\bibitem[{Zhou et~al.(2023{\natexlab{a}})Zhou, Lu, Mishra, Brahma, Basu, Luan, Zhou, and Hou}]{zhou2023instruction}
Jeffrey Zhou, Tianjian Lu, Swaroop Mishra, Siddhartha Brahma, Sujoy Basu, Yi~Luan, Denny Zhou, and Le~Hou. 2023{\natexlab{a}}.
\newblock Instruction-following evaluation for large language models.
\newblock \emph{arXiv preprint arXiv:2311.07911}.

\bibitem[{Zhou et~al.(2023{\natexlab{b}})Zhou, Jiang, Wilcox, Cotterell, and Sachan}]{zhou2023controlled}
Wangchunshu Zhou, Yuchen~Eleanor Jiang, Ethan Wilcox, Ryan Cotterell, and Mrinmaya Sachan. 2023{\natexlab{b}}.
\newblock Controlled text generation with natural language instructions.
\newblock In \emph{International Conference on Machine Learning}, pages 42602--42613. PMLR.

\bibitem[{Zhou et~al.(2023{\natexlab{c}})Zhou, Liu, Yang, Shao, Liu, Yue, Ouyang, and Qiao}]{zhou2023beyond}
Zhanhui Zhou, Jie Liu, Chao Yang, Jing Shao, Yu~Liu, Xiangyu Yue, Wanli Ouyang, and Yu~Qiao. 2023{\natexlab{c}}.
\newblock Beyond one-preference-for-all: Multi-objective direct preference optimization.
\newblock \emph{arXiv preprint arXiv:2310.03708}.

\end{thebibliography}

\appendix

\newpage
\clearpage

\section{Hard Attributes}
\label{sec:hard_attr}

\begin{table*}[htbp]
\centering
\small
\begin{tabular}{l|p{2.6cm}|p{9cm}}
\toprule
\textbf{Instruction Group} & \textbf{Instruction} & \textbf{Description} \\
\midrule
Keywords & Include Keywords & Include keywords \{keyword1\} in your response \\
Keywords & Keyword Frequency & In your response, the word {word} should appear \{N\} times. \\
\midrule
Length Constraints & Number Paragraphs & Your response should contain \{N\} paragraphs. You separate paragraphs using \textbackslash n \textbackslash n \\
Length Constraints & Number Words & Answer with at least / around / at most \{N\} words. \\
Length Constraints & Number Sentences & Answer with at least / around / at most \{N\} sentences. \\
\midrule
Change Cases & All Uppercase & Your entire response should be in English, capital letters only. \\
Change Cases & All Lowercase & Your entire response should be in English, and in all lowercase letters. No capital letters are allowed. \\
\midrule
Start with & Start With & Finish your response with this exact phrase \{end\_phrase\}. No other words should follow this phrase. \\
\bottomrule
\end{tabular}
\caption{The list of 8 verifiable instructions, with brief descriptions. We use these instructions because we think they are either easy to verify or common in real-world applications.}
\label{tab:list-of-verifiable-instruction}
\end{table*}

The hard attributes employed in this study, as detailed in Table \ref{tab:list-of-verifiable-instruction}, comprise a set of verifiable instructions designed to enforce precise, programmatically assessable constraints on text generation. These attributes are categorized into four primary groups: (1) Keywords, which mandate the inclusion or frequency of specific terms (e.g., "Include \{keyword1\}" or "appear \{N\} times"); (2) Length Constraints, governing structural requirements such as paragraph count, word limits, or sentence boundaries; (3) Change Cases, enforcing syntactic rules like all-uppercase or all-lowercase formatting; and (4) Positional Directives, such as starting responses with predefined phrases. Each attribute is selected for its objective verifiability through rule-based checks while also reflecting common real-world application scenarios, such as compliance with stylistic guidelines or technical specifications. By anchoring the evaluation in these deterministic constraints, the framework guarantees rigorous assessment of model adherence to fine-grained requirements, aligning with the dataset's emphasis on combinatorial complexity and practical utility.

\section{Generalization to Unseen Attributes}
We evaluate the models’ ability to generalize to unseen, more challenging attributes, focusing on two types:
\begin{enumerate} \item \textbf{Absolute Position of a Word}: The k-th (k $\le$ 5) word in the text must be A.
\item \textbf{Relative Position Between Two Words}: Word A must appear before word B.
\end{enumerate}
We use text from the FineWeb validation set and extract 50 attributes per document, focusing on these two types of harder attributes. We then evaluate the three models on this benchmark. The results show that the original model achieves a score of 21.56, while auto-reconstruct slightly reduces performance to 20.79. However, incorporating GPO alongside AR improves generalization, yielding a score of 24.05, suggesting that GPO enhances the model’s ability to handle these harder constraints.


\section{Dataset Statistics}
\label{appendix:statistics}
\begin{table}[htbp]
\centering
\resizebox{0.49\textwidth}{!}{
    \begin{tabular}{lcccccc}
    \hline
    \textbf{Subset} & \textbf{Train Size} & \textbf{Val Size} & \textbf{Avg Length} & \textbf{Soft Attrs} & \textbf{Hard Attrs} & \textbf{Total Attrs} \\
     & & & (words) & (per sample) & (per sample) & (per sample) \\
    \hline
    FineWeb (Local) & 6,159 & 200 & 361.6 & 7.80 & 38.10 & 45.90  \\
    Multi-sources (Global) & 1600 & 400 & - & 5.1 & 24.8 & 29.9 \\
    \hline
    \end{tabular}
}
\caption{UltraBench Dataset Composition}
\label{tab:overall_statistic}
\end{table}
In this section, we present detailed statistics of our dataset, including a comparison with existing datasets, quality control evaluation, and the composition of our multi-sources subset.

\paragraph{Comparison with Existing Datasets.}
\newcolumntype{g}{>{\columncolor{green!10}}c}
\setlength\tabcolsep{7pt}
\begin{table}[htbp]
\centering
\huge
\newcolumntype{b}{>{\columncolor{blue!10}}c}
\renewcommand{\arraystretch}{1.6}
\resizebox{0.5\textwidth}{!}{

\begin{tabular}{lccccc}

\toprule
\multicolumn{1}{c}{\multirow{2}{*}{Method}} & \multicolumn{4}{c}{Data Quality} &  \\ \cline{2-6} 
\multicolumn{1}{c}{}                       & Nums.       & Cons.    & Avg Attr.      & Synt.    \\ \midrule
IFeval~\cite{zhou2023instruction} & 541  & H & 1.54 & \ding{51} \\
FollowBench~\cite{jiang2023followbench} & 820 & H/S & 3.0 & \ding{51}  \\
CFBench~\cite{zhang2024cfbench} & 1000 & H/S & 4.24 & \ding{55} \\
InFoBench~\cite{qin2024infobench} & 500 & H/S & 4.5 & \ding{55} \\
\our (FineWeb Split) & 6159 & H/S & \textbf{45.9} & \ding{55} \\
\our (Multi-source Split) & 1600 & H/S & \textbf{29.9} & \ding{55} \\
\bottomrule
\end{tabular}%
}
\caption{
  Detailed comparison of relevant works. Ours
represents our dataset construction approach. \textquotesingle Nums.\textquotesingle, \textquotesingle Cons.\textquotesingle, \textquotesingle Avg Attr.\textquotesingle,
and \textquotesingle Synt.\textquotesingle\  denote the number of samples, constraint types, average number of attributes, and whether the data is synthesized.
}

  \label{tab:comparison}
\end{table}

Table~\ref{tab:comparison} provides a comparison between our dataset and several representative constraint-based datasets, including IFeval~\cite{zhou2023instruction}, FollowBench~\cite{jiang2023followbench}, CFBench~\cite{zhang2024cfbench}, and InFoBench~\cite{qin2024infobench}. Our dataset distinguishes itself with a significantly larger number of samples (6,159) and a notably higher average number of attributes per instance (45.9). Unlike prior datasets, which primarily rely on either human annotations or simple constraints, our data features a rich combination of both hard and soft constraints, offering a more challenging and comprehensive benchmark. Importantly, our data is not synthesized, ensuring its alignment with real-world use cases.

\paragraph{Domain Composition in Multi-sources Subset}
\begin{table}[htbp]
\centering
\resizebox{0.5\textwidth}{!}{
\begin{tabular}{lcc}
\hline
\textbf{Category} & \textbf{Domain} & \textbf{Percentage} \\
\hline
\multirow{4}{*}{Web Data} & CC News (Middle) & 9.24\% \\
& Falcon RefinedWeb Filtered & 9.59\% \\
& CC EN (Middle) & 9.86\% \\
& C4 Filtered & 10.03\% \\
\hline
\multirow{2}{*}{Forum} & Reddit & 9.75\% \\
& StackExchange (RedPajama) & 10.42\% \\
\hline
\multirow{2}{*}{Papers} & arXiv (RedPajama) & 10.05\% \\
& Pile-Extracted Scientific Open (PES2O) & 9.79\% \\
\hline
Books & Books & 10.42\% \\
Wikipedia & Wikipedia & 10.84\% \\
\hline
\end{tabular}
}
\caption{Distribution of data sources in the multi-sources subset by category. The data sources are grouped into major categories: Web Data, Forums, Papers, Books, and Wikipedia. Percentages represent the proportion of each domain within the subset.}
\label{tab:rl_domains}
\end{table}

\begin{figure}[htbp]
    \centering
        \includegraphics[width=.5\textwidth]{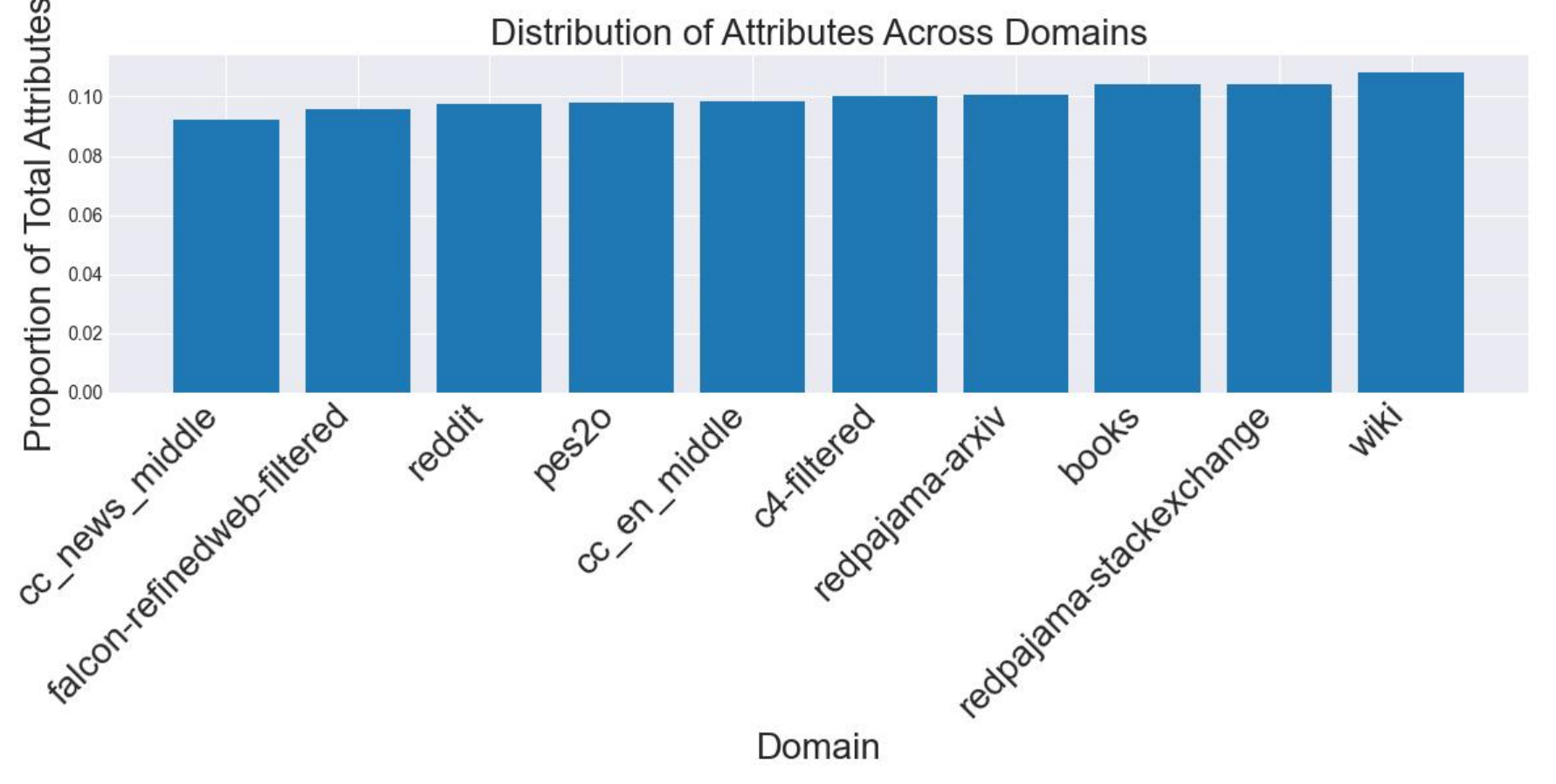}
    \caption{Proportion of Attributes Across Different Data Domains. The bar chart visualizes the relative contribution of each domain to the multi-sources subset, highlighting a balanced distribution across various sources such as web data, forums, papers, books, and Wikipedia.}
    \vspace{-1em}
    \label{fig:rl_distribution}
\end{figure}
Our multi-sources subset is constructed from a diverse range of data sources, encompassing web data, forums, academic papers, books, and Wikipedia. Figure~\ref{fig:rl_distribution} illustrates the proportion of attributes contributed by each domain, highlighting a balanced distribution across these categories. Table~\ref{tab:rl_domains} further details the exact composition, showing that no single source overwhelmingly dominates, ensuring robustness and variety in downstream tasks.

\paragraph{Quality Control Metrics}
Maintaining data quality is critical for ensuring reliable evaluations. We assess the agreement rate (AR) between human annotators and the final dataset as a key metric. As summarized in Table~\ref{tab:quality_metrics}, the FineWeb subset achieves an AR of 92.3\%, while the multi-sources subset attains 88.7\%. These high agreement rates reflect the robustness of our data curation process, confirming that both subsets align closely with human judgment.
\begin{table}[htbp]
\centering
\resizebox{0.5\textwidth}{!}{
\begin{tabular}{lcc} 
\toprule
\textbf{Metric} & \textbf{FineWeb Subset} & \textbf{Multi-sources Subset} \\
\midrule
Agreement Rate (AR) & 97\% & 96\% \\
\bottomrule
\end{tabular}
}
\caption{Quality Control Metrics}
\label{tab:quality_metrics}
\end{table}

\section{DPO data quality} 
\begin{table}[htbp]
    \centering
    \small
    \begin{tabular}{p{7cm}}
     \toprule
\textbf{High Correlation: } \\
\quad - Thought-provoking narrative with a call to action \\
\quad - Author's Name and Location Identifier: The text begins with the name S. TEITELBAUM followed by a location ST. JOHNS, FL. \\
\quad - Engaging Headline: The title captures the reader's attention by listing 5 Reasons for a specific action. \\
\textbf{Low Correlation: } \\
\quad - AI Leadership: The partnership aims to position Singapore as a leader in AI within healthcare \\
\quad - Focus on Competitive Standards: The passage stresses the competitiveness of FAU’s admissions process \\
\bottomrule
    \end{tabular}
    \caption{Correlation Examples}
    \label{tab:high_correlation_examples}
\end{table}
In this section, we showcase some examples sampled by our global selection strategy.
\paragraph{High Correlation} 
Our attribute correlation modeling step aims to select semantically coherent and mutually reinforcing attributes during GPO training. This process effectively groups attributes that frequently co-occur in natural text, leading to the selection of high-quality attribute combinations. 

\paragraph{Low Similarity}
\begin{table}[htbp]
    \centering
    \small
    \begin{tabular}{p{7cm}}
     \toprule
\textbf{Low Similarity: } \\
\quad - Focus on Natural Ingredients: Emphasizes the importance of natural ingredients \\
\quad - Protein-rich for Satiety and Muscle Growth: The high protein content in buffalo milk helps increase satiety \\
\textbf{High Similarity: } \\
\quad - Health focus: The text emphasizes overall health benefits\\
\quad - Detailed explanation for each benefit: Each health benefit mentioned is followed by an explanation or reasoning \\
\bottomrule
    \end{tabular}
    \caption{Similarity Examples}
    \label{tab:similarity_examples}
\end{table}
While high correlation ensures that attributes are semantically aligned, it is equally important to maintain attribute diversity to prevent redundancy and overfitting. Our global selection strategy aims to minimize the presence of highly similar attributes within the same prompt. For instance, attributes like \emph{“Engaging Headline”} and \emph{“Attention-Grabbing Title”} convey nearly identical meanings and offer little additional training value when paired together. By prioritizing low-similarity combinations, we encourage the model to generalize across a broader range of attribute expressions, improving its adaptability to diverse prompts.

\begin{figure}[htbp] 
    \centering
        \includegraphics[width=0.5\textwidth]{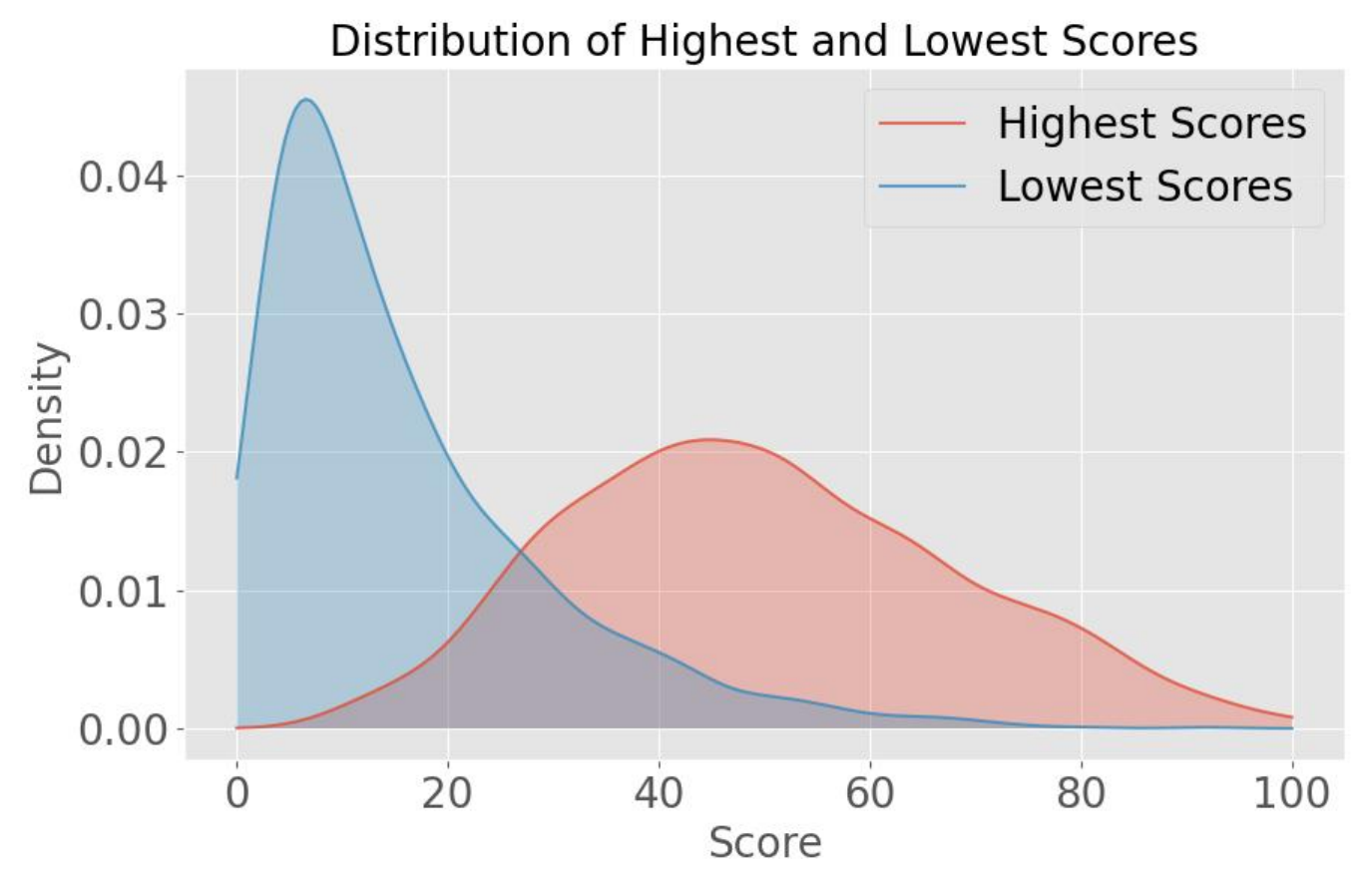}
    \caption{Score distributions of chosen and rejected data.}
    \label{fig:dpo_score_distribution}
\end{figure}
To further illustrate the effectiveness of our sampling strategy in the GPO stage, we present several representative cases selected by our attribute-based sampling approach. These examples demonstrate the diversity and coverage achieved through our strategy, highlighting both common and edge-case attribute combinations. 

\section{Prompts Used in This Study}
We employ three distinct prompts to support different stages of our EFCG pipeline: Decomposition, Judging, and Generation.

\paragraph{Decomposition Instruction.}
\begin{table}[htbp]
    \vspace{-0.5em}
    \centering
    \small
    \begin{tabular}{p{7cm}}
     \toprule
\#\#\# Requirements \\
For the following paragraph, propose attributes that capture its overall characteristics. Focus on what makes this text unique and distinctive, rather than using predefined categories. Your analysis should: \\
- Identify the most prominent and defining features of the text \\ 
- Use clear, specific descriptions rather than vague terms \\
- Base attributes solely on what is explicitly present in the text \\
- Describe each attribute with enough detail to be meaningful \\
Avoid: \\
- Overly broad or generic attributes \\
- Speculative interpretations \\ 
- Attributes not clearly supported by the text \\ 
- Complex or academic jargon \\

Output each attribute on a separate line, separated by a single newline, with no line breaks within each attribute. \\

Now, analyze the following paragraph and summarize its key attributes: \\

\#\#\# Text \\
\{text\}

\#\#\# Attributes  \\
\bottomrule
    \end{tabular}
    \caption{Decompose Prompt}
    \label{tab:decompose_prompt}
\end{table}
This prompt is used to extract a set of soft attributes from a given text. The goal is to decompose the paragraph into its most defining characteristics, capturing both stylistic and semantic elements. Models are instructed to focus on identifying specific, explicit features of the text rather than relying on generic descriptions or subjective interpretations. Attributes must reflect the unique aspects of the text and be grounded in the content. 

\paragraph{Judge Instruction.}
\begin{table}[htbp]
    \centering
    \small
    \begin{tabular}{p{7cm}}
     \toprule
You are a binary evaluator. Given a text and several attributes, determine if the text fulfills each attribute. \\

Your task is simple: \\
- Score 0 if the text does NOT fulfill the attribute or the attribute is not directly mentioned \\
- Score 1 if and only if the text directly fulfills the attribute \\

Text to evaluate: \\
\{text\} \\

Attributes to evaluate: \\
\{attributes\} \\

Provide exactly \{num\_attributes\} scores, one per line, using this format: \\
Score: 0 or 1 \\

- Scores should correspond to attributes in order \\
- Only provide the scores, no additional explanation \\
\bottomrule
    \end{tabular}
    \caption{Judge Prompt}
    \label{tab:judge_prompt}
\end{table}
This prompt serves as a binary evaluation guideline to determine whether a generated text satisfies a given set of attributes. Evaluators are asked to assess each attribute independently, assigning a score of 1 if the text explicitly fulfills the attribute and 0 otherwise. The evaluation is strict, requiring the text to directly align with the specified attribute for a positive score.

\paragraph{Generation Instruction.}
\begin{table}[htbp]
    \centering
    \small
    \begin{tabular}{p{7cm}}
     \toprule
You are an expert at generating text that matches given attributes. Your task is to generate a text that satisfies as many of the provided attributes as possible. \\

\#\#\# Hard Attributes: \\
\{hard\_attributes\} \\

\#\#\# Soft Attributes: \\
\{soft\_attributes\} \\
\bottomrule
    \end{tabular}
    \caption{Generation Prompt}
    \label{tab:generation_prompt}
\end{table}
This prompt is used to instruct the language model to generate a piece of text that aligns with a provided set of hard constraints and soft attributes. Hard attributes typically represent structural or factual constraints (e.g., budget, schedule), while soft attributes reflect stylistic or semantic preferences (e.g., tone, vividness). The model is guided to generate text that adheres to as many of these attributes as possible, balancing the satisfaction of both hard and soft constraints.


\section{The Complete Case Study}
\label{appendix:case_study}

\begin{table*}[htbp]
    \centering
    \small
    \begin{tabular}{p{14cm}}
     \toprule
\#\#\#  Objective: \\
Generate a 5-day family travel itinerantry that satisfies all specified requirements while adhering to highly fine-grained constraints. The generated itinerary should balance real-time adaptability, strict hard attributes, and semantic soft attributes. \\

\#\#\# User Profile: \\
 - Travelers: 2 adults + 1 child (age 8) \\
 - Budget: $<=$ \$300/day (total \$1,500 for the trip) \\
 - Activity Balance: 70\% educational/cultural experiences, 20\% relaxation, 10\% family-friendly shopping. \\

\#\#\# Hard Attributes: \\
- Activity Scheduling: \\
\quad- Each activity must have a defined start and end time, ensuring there is no overlap between activities. \\
\quad- A break period from 13:00-14:30 is mandatory daily. \\
\quad- Each activity must fit within a 2-hour window unless otherwise specified. \\

- Budget Requirements: \\
\quad- Each day’s total cost (including transportation, food, and activities) must not exceed \$300. \\
\quad- Transportation is limited to metro and walking only, with a maximum of 3 metro rides per day. \\

- Location Constraints: \\
\quad- Must-visit locations: City Zoo (Day 1) and Science Museum (Day 3). \\
\quad- Activities must occur in geographically adjacent areas to minimize walking distance. \\

- Keyword Requirements: \\
\quad- Each day’s description must include specific keywords. For example: \\
\quad- Day 1: “wildlife,” “exploration,” and “interactive learning.” \\
\quad- Day 3: “science,” “innovation,” and “hands-on exhibits.” \\

- Structure Constraints: \\
\quad- Each day’s itinerary must consist of 4 sections: \\
\quad\quad- Morning activity \\
\quad\quad- Break/lunch period \\ 
\quad\quad- Afternoon activity \\
\quad\quad- Evening summary (limited to 50 words) \\

\#\#\# Soft Attributes \\
- Tone and Emotion: \\ 
\quad- Day 1: Use a tone that conveys “excitement and discovery.” \\ 
\quad- Day 3: Use a tone that conveys “curiosity and wonder.” \\
- Language Style: \\ 
\quad- Use descriptive, vivid, and family-friendly language throughout. \\
\quad- Include at least one metaphor or simile per day (e.g., "The Science Museum felt like stepping into the future!"). \\
- Visual Details: \\
\quad- Each activity must include specific sensory details (e.g., "the bright colors of the parrots at the zoo" or "the tinkling sound of water fountains at the park").

- Adaptive Adjustments (Real-time Constraints): \\
\quad- Weather Sensitivity: \\
\quad\quad- If the rain forecast exceeds 60\%, replace outdoor activities with indoor alternatives while keeping the overall tone and keywords intact. \\ 
\quad- Physical Endurance: \\
\quad\quad- If a day’s total walking distance exceeds 10 kilometers, the next day’s activities must reduce walking by 30\%. \\
\quad- Health Responsiveness: \\
\quad\quad- If a health-related issue arises (e.g., fatigue or illness), adjust the itinerary dynamically to: \\
\quad\quad- Reduce activity duration to half. \\ 
\quad\quad- Substitute the activity with a more relaxing or passive option. \\
\bottomrule
    \end{tabular}
    \caption{The complete travel planner case study.}
    \label{tab:travel_planner_case}
\end{table*}
The travel planner case study exemplifies the practical usefulness of EFCG in handling complex, multi-faceted requirements. As shown in Table~\ref{tab:travel_planner_case}, generating a 5-day travel itinerary involves satisfying a diverse set of hard attributes (e.g., budget limits, time scheduling, location constraints) alongside soft attributes (e.g., tone, emotion, visual details), while also adapting to real-time factors like weather and physical endurance. Such a task necessitates precise control over both hard and soft constraints, making it a natural testbed for evaluating EFCG systems.

\end{document}